# Optimizing Large Language Models for Detecting Symptoms of Comorbid Depression or Anxiety in Chronic Diseases: Insights from Patient Messages

Jiyeong Kim[1], PhD, MPH, Stephen P. Ma[2,3,4], MD, PhD, Michael L. Chen[1], BA, Isaac R. Galatzer-Levy[5], PhD, John Torous[6], MD, MBI, Peter J. van Roessel[7], MD, PhD, Christopher Sharp[2,3,4], MD, Michael A. Pfeffer[2,3,4], MD, Carolyn I. Rodriguez[7,8]*, MD, PhD, Eleni Linos[1]*, MD, DrPH, Jonathan H. Chen[2,9]*, MD, PhD

*Joint senior authors

[1] Stanford Center for Digital Health, Department of Medicine, Stanford University, Stanford, CA, USA
[2] Division of Hospital Medicine, Stanford School of Medicine, Stanford, CA
[3] Technology and Digital Solutions, School of Medicine, Stanford University, Stanford, CA, USA
[4] Stanford Health Care, Palo Alto, CA, USA
[5] Google LLC, Mountain View, CA, USA
[6] Division of Digital Psychiatry, Department of Psychiatry, Beth Israel Deaconess Medical Center, Boston, MA, USA
[7] Department of Psychiatry and Behavioral Sciences, School of Medicine, Stanford University, Palo Alto, CA, USA
[8] Veterans Affairs Palo Alto Health Care System, Palo Alto, CA, USA
[9] Department of Biomedical Data Science, School of Medicine, Stanford University, Stanford, CA, USA

**Abstract**

Patients with diabetes are at increased risk of comorbid depression or anxiety, complicating their management. This study evaluated the performance of large language models (LLMs) in detecting these symptoms from secure patient messages. We applied multiple approaches, including engineered prompts, systemic persona, temperature adjustments, and zero-shot and few-shot learning, to identify the best-performing model and enhance performance. Three out of five LLMs demonstrated excellent performance (over 90% of F-1 and accuracy), with Llama 3.1 405B achieving 93% in both F-1 and accuracy using a zero-shot approach. While LLMs showed promise in binary classification and handling complex metrics like Patient Health Questionnaire-4, inconsistencies in challenging cases warrant further real-life assessment. The findings highlight the potential of LLMs to assist in timely screening and referrals, providing valuable empirical knowledge for real-world triage systems that could improve mental health care for patients with chronic diseases.

**[142/150 words]**

**Main**
Patients with diabetes are more than twice as likely to suffer from comorbid depression or anxiety,[1,2] increasing their risk of hospitalization, functional disability, complications, and mortality.[3] The bi-directional relationship between diabetes and comorbid depression can exacerbate both conditions.[4] Detecting symptoms of comorbid depression or anxiety is challenging due to patient unawareness or reluctance to report symptoms, clinicians' limited time and expertise, and overlapping symptoms. Additionally, a shortage of trained psychiatrists and collaborative care teams further complicates timely diagnosis.[5,6] Addressing these issues is crucial to prevent disease progression and manage both physiological and psychiatric aspects effectively.

Patient-centered care, which includes addressing concerns, providing guidance, and involving patients in decision-making, is crucial for patients with depression or anxiety.[7] A patient-centered digital health platform enhances engagement, self-management, and treatment adherence.[8] Secure messaging via patient portals allows patients to communicate directly with clinicians, aiding medical management.[9] Patients with diabetes are among the most engaged user groups with the patient portal.[8] Therefore, secure patient messages are an important resource for improving patient-centered care beyond traditional clinical notes.[10]

Large language models (LLMs) have shown clinician-comparable levels of medical knowledge and competitive diagnostic performance in various medical specialties.[11,12] In psychiatry, LLMs outperformed mental health professionals in detecting obsessive-compulsive disorder.[13] Similarly, LLMs were promising in predicting the clinical prognosis of schizophrenia, assisting in triage recommendations for mental health referrals, and detecting the risk of depression through user-generated text data.[14–16] Though machine learning models have shown good performance, translating patients' self-reported history into specific psychiatric symptoms and ultimately diagnoses has been challenging.[17,18] Currently, there is a critical knowledge gap in the performance of LLMs in detecting depressive or anxious symptoms, specifically in the context of secure patient messages, despite the importance and the potential promise of LLMs.

Thus, we aim to comprehensively evaluate the performance of LLMs in detecting symptoms of comorbid depression or anxiety from secure patient messages. To identify the best-performing model in this context and to specify the effective strategies to enhance the models' performance, we applied multiple approaches, including engineered prompts for multi-level tasks, systemic persona, temperature adjustments, and zero-shot and few-shot learning. These findings could serve as empirical evidence of LLMs' use for detecting symptoms of comorbid depression or anxiety from secure patient messages, highlighting the key factors to consider for optimal performance.

**Results**

***Performance of LLMs: zero-shot vs few-shot***
The performance of LLMs in detecting depressive or anxious symptoms from patient messages differed by model and approaches (Figure 2). In zero-shot settings, Llama 3.1 405B showed the highest performance in F-1 (0.93 [95% CI 0.91-0.95]), recall (0.95 [0.93-0.97]), and accuracy (0.93 [0.91-0.95]), and Llama 3.1 8B and DeepSeek R1 also showed excellent F-1 (0.91 [0.88-0.93] and 0.90 [0.88-0.93]) and accuracy (0.90 [0.88-0.93] and 0.91 [0.89-0.93]), respectively (Figure 2. a). Two reasoning models showed nearly perfect precision (OpenAI o1: 1.00 [0.99-1.00]; DeepSeek R1: 0.99 [0.97-1.00]) yet presented much different recalls (OpenAI o1: 0.68 [0.62-0.73]; DeepSeek R1: 0.84 [0.79-0.88]).

Across three knowledge models (Llama 3.1 8B and 405B, and Gemini Pro 1.5), few-shot learning increased precision while decreasing recall (Figure 2. b and c). Few-shot learning increased the performance of Gemini Pro 1.5 compared to zero-shot, including F-1 (0.87→0.89, *n*=12), yet it did not effectively enhance other models' accuracy and F-1 score. All results of zero-shot vs few-shot are available in **Appendix 5**.

### *Performance of LLM: PHQ-4 vs binary classification*
While the performance of binary classification was superior in all tested LLMs, three knowledge models showed overall competitive F-1 scores and accuracy (over 0.8) when using PHQ-4 compared to using binary classification in zero-shot settings in detecting depressive or anxious symptoms from patient messages: F-1 (0.82 [0.78-0.85] Llama 3.1 8B; 0.87 [0.84-0.89] Llama 3.1 405B; 0.83 [0.80-0.86] Gemini Pro 1.5) (Figure 3). However, two reasoning models' F-1 scores and accuracy were significantly poorer when PHQ-4 was used compared to binary-based symptom detection [PHQ-4 vs binary]: OpenAI o1, F-1 [0.61 vs. 0.81], *p*<0.0001; DeepSeek R1, F-1 [0.51 vs. 0.90], *p*<0.0001.

### *Effect of systemic persona*

#### Effect of persona on few-shot learning
In zero-shot, F-1 and accuracy decreased with systemic persona compared to baseline in Llama 3.1 8B (F-1: 0.91→ 0.88), Llama 3.1 405B (F-1: 0.93→0.87), and DeepSeek R1 (F-1: 0.90→0.87), yet those increased with persona in Gemini Pro 1.5 (F-1: 0.87→0.90, accuracy: 0.88→ 0.90) (Figure 4). However, in few-shot settings, the systemic persona mostly enhanced F-1 and accuracy consistently across models up to 0.09 (e.g. F-1 0.73→0.82, Gemini Pro 1.5). All results of the effect of systemic persona are available in **Appendix 6**.

#### Effect of persona on mental health measurement metric
Applying systemic persona increased precision of LLMs, however, it decreased F-1, recall, and accuracy in PHQ-4 based classification, widening the gap with binary classification.

### *Performance of LLMs in challenging cases*
In challenging cases, DeepSeek R1 showed a competitive F-1 score (0.84 [0.67-0.97]) and accuracy (0.87 [0.74-0.97]) (Figure 5). The remaining five models—including two knowledge models (Llama 3.1 8B and 405B) that showed the highest F-1 and accuracy in our initial assessments and the three latest reasoning models (OpenAI o3-mini, OpenAI o1 and Gemini Pro 2.0 Thinking) —performed much worse (F-1 scores of 0.52-0.70). Unlike the results of our initial assessments, LLMs showed high recall (over 0.9 in four LLMs) but low precision (0.43-0.58 in four LLMs) in challenging cases. Two psychiatrists agreed that the LLM reasoning was logical, and the explanations were thorough.

**Discussion**
We comprehensively assessed LLMs' performance in detecting symptoms of depression or anxiety from secure patient messages of those with diabetes. We found that three out of five LLMs showed excellent performance with over 90% in both F-1 and accuracy. Llama 3.1 405B was the highest performing model (both F-1 and accuracy of 93%) with a zero-shot approach. Moreover, the two reasoning models (OpenAI o1 and DeepSeek R1) presented nearly perfect precision, highlighting their potential usefulness even in the resource-constrained health system. The LLMs not only showed a promising capacity to detect anxiety and depression symptoms in a binary classification but could also handle a complex screening metric (PHQ-4). Finally, for the challenging cases, only one reasoning model (DeepSeek R1) presented a competitive F-1 score (84%) and accuracy (87%). The findings from our comprehensive LLM evaluations of clinical messages from patients highlight the significant potential of LLMs in detecting symptoms of comorbid depression and anxiety, which could be considered as assistance for timely screening and referrals.

The excellent F-1 score and accuracy with a zero-shot approach highlight the efficiency of the application of LLM in detecting symptoms of comorbid depression or anxiety without fine-tuning or further tailored approaches. Previously, LLMs with zero-shot classification showed limited capacity in mental health prediction tasks, suggesting recent improvements of these models.[25] Given that those tailoring efforts could increase the cost of LLM use in healthcare and frequently require significant expertise that is not always widely available, the superlative performance of the zero-shot approach was particularly encouraging.[26] Moreover, while previous studies using AI/LLMs heavily focused on vignette-based diagnostic performance, our findings are applicable to real-world screening scenarios.[27,28] The LLMs' ability to assist in screening depression or anxiety through text-based communications could open novel opportunities for timely detection of symptoms of comorbid depression or anxiety among those with various chronic diseases.

However, despite the LLMs' excellent performance, immediate clinical applications require caution due to the observed inconsistencies in challenging cases. As challenging cases were defined as messages that contained words that expressed negative emotions but were not determined as depressive or anxious by two highly experienced psychiatrists, these cases may be prevalent in actual clinical settings. Hence, further assessment is warranted to see how LLMs perform in real-life settings. In the current study, LLMs' reasoning in challenging cases was comprehensive, which aligned with previous findings,[29] offering cautious optimism about the potential for LLMs to augment clinicians' workflow. Piloting text-based mental health screening in clinical settings is a suggested next step to evaluate the practical impact of this approach, including symptom detection rate, operational cost, care efficiency, clinicians' acceptance, and patients' perspectives. These rigorous evaluations will help guide the use of LLMs as an automated screening agent to assist clinicians at the front line for enhanced mental health care.

We observed robust zero-shot accuracy (85% and higher in three LLMs) using multi-class and multi-level symptom detection for depression or anxiety (PHQ-4), unlike previously reported poor performance (39.6-65.6%) in four-level classification using social media data.[25] Our findings of LLMs' capacity to process the complex mental health measurement tool at a promising level highlight the enhanced explainability of LLMs' performance in text-based symptom detection for comorbid depression or anxiety. By successfully classifying each message into four different domains (e.g. worrying, little interest, hopeless, and nervous) and rating the severity level for each domain as directed, LLMs demonstrated that they had the ability of comprehensive assessment, and the multi-step process was interpretable. Notably, all knowledge models (F-1 and accuracy) outperformed reasoning models with the PHQ-4 based assessment, suggesting that the knowledge models are more suitable for this type of symptom detection task. In this study, the performance of binary classification was better than PHQ-4 based assessments. The use of binary classification in preparing the reference data may have influenced the outcomes, warranting further investigation in the future.

Proper diagnosis is especially challenging because comorbid depression or anxiety shares similar symptomatic issues with some chronic conditions, including diabetes (e.g. loss of energy, trouble sleeping). Unnecessary false positive alerts may discourage the adoption of LLM use among clinicians due to anticipated burden.[30] Hence, models with high precision could be optimal. While two reasoning models showed perfect precision in zero-shot (99-100%), all three knowledge models (including Llama 3.1 8B) also easily achieved 99% precision with 2-4 few-shot learning. As cost is an important factor for LLMs in healthcare, the ability to achieve comparable precision when necessary using small, open-source local models is critical for widespread adoption.

This study has limitations. First, the message samples were from a single academic medical center, which could limit the generalizability of findings. However, the message data included messages from 22 affiliated centers in California. Moreover, the benchmark patient messages are more than 600, which may have included a diverse range of conversational topics such as messages sent by the caretakers. Despite the limitations, the findings from this comprehensive evaluation of various LLMs contribute valuable insights to the field. By utilizing secure patient messages, we demonstrated the performance of LLMs in real-world applications relying on our patients' own words. Moreover, we evaluated multiple new state-of-the-art reasoning models, which have not yet been assessed in the healthcare domain.

Given that only 5% of LLMs have been tested with patient data[31], our findings on LLM performance in detecting symptoms of comorbid depression or anxiety through secure messages from individuals with diabetes offer significant empirical knowledge. The novel knowledge could serve as a foundation for implementing real-world triage and screening systems, which could initiate timely treatment and care, eventually enhancing the health outcomes in millions of patients.

## Methods

### Data source and study design
We obtained secure patient messages received from individuals with diabetes (ICD-10 codes: E08, E09, E10, E11, E13) as well as depression (ICD-10 code: F32, F33) or anxiety (ICD-10 code: F41) through the secure patient portal of a large academic medical center (Stanford Health Care, [SHC]) and 22 affiliated centers in California in 2013-2024. We included clinical issues labeled as patient medical advice request (PMAR) routed to internal medicine, family medicine, and primary care clinics. The Stanford Institutional Review Board approved this study.

### Preparation of benchmark data
The secure patient messages were deidentified using the Safe Harbor method. Before inputting messages into the LLM, two researchers (JK and CIR) reviewed each message to ensure that no protected health information was included in our data set. We collected the most recent six-month messages (10/2023-04/2024) and randomly ordered them.

For positive benchmark (Yes [1], depression or anxiety symptoms are present), we used messages from patients *with diabetes and comorbid depression or anxiety* (Figure 1). The first researcher (JK) labeled messages containing at least one depression or anxiety-related keyword, and negative sentiment as 'Yes $[1]_{preliminary}$'. An experienced psychiatrist (CIR) reviewed and confirmed these as 'Yes $[1]_{confirmed}$' if the patient seemed to need further assessment for depression or anxiety. For negative benchmark (No [0], depression or anxiety symptoms are not present), we used messages from patients *with diabetes but no comorbid depression or anxiety.* JK labeled messages without a depression or anxiety-related keyword and with neutral or positive sentiment as 'No $[0]_{preliminary}$'. CIR confirmed these as 'No $[0]_{confirmed}$' if no further assessment was needed. A total of 606 reference messages were prepared (No [0], *n*=301; Yes [1], *n*=305).

### Identification of patients' language for depression or anxiety
To identify patients' language that they used to describe their depression or anxiety symptoms via secure messaging, we analyzed the messages from those with diabetes and comorbid depression or anxiety (Figure 1). By leveraging Bidirectional Encoder Representations from Transformers (BERT) and Balanced Iterative Reducing and Clustering (BIRCH) algorithms, we developed a 2-staged natural language processing (NLP) topic model.[19,20] In the first topic model, similar messages were clustered by cosine similarity score, which we set as 0.82, and key topic groups were created. In the second topic model, the generated topic groups were clustered once again by similarity, keeping the essential topics. The topic model assigned representative keywords of each topic. Through the topic modeling, we obtained keywords that were grouped together with depression or anxiety (e.g. mood, worry, panic, stress), which were used for benchmark data preparation. A full list of keywords is in **Appendix 1**.

### Evaluation of LLMs
The primary outcome was the F-1 score, precision, recall, and accuracy computed by the binary classification of depression or anxiety symptoms (Yes [1] vs No [0]) using the benchmark data. Using a bootstrapping method (*n*=1,000), we computed 95% confidence intervals of each primary outcome. Given that the patient messages offered limited information about the patient's history, we did not precisely separate depression and anxiety symptoms, instead combining the symptoms as one outcome.

We carefully selected five LLMs for evaluation—three knowledge models by size: 1) small: Llama 3.1 8B (Meta Inc., July 2024), 2) medium: Gemini Pro 1.5 (Google LLC., September

2024), 3) large: Llama 3.1 405B (Meta Inc., July 2024 ); and two reasoning models: 1) OpenAI o1 (OpenAI Inc., September 2024), 2) DeepSeek R1 (January 2025)—to comprehensively compare the performance and the changes of the performance by approach. Applying various strategies, we intended to identify the optimal approaches by model: 1) zero-shot vs few-shot ($n$=2, 4, 6, 8,10, and 12); 2) binary classification vs Patient Health Questionnaire (PHQ)-4 based classification; 3) with vs without systemic persona; and 4) temperature of the test environment (0.6 vs 0.3). We performed all of the assessments in secure analytics environments provided by SHC that ensure data privacy and security through private API endpoints and end-to-end encryption.[21] All analytic codes and prompts used in this study are available in **Appendix 2**.

**Zero-shot vs few-shot learning**. We applied a few-shot learning with reference examples of how to diagnose the message annotated with a highly experienced psychiatrist's (CIR) reasoning ($n$=2, 4, 6, 8, 10, and 12). Each set of reference examples included a balanced sample of positive (Yes [1]) and negative (No [0]) references.

**Binary classification vs PHQ-4-based classification**. While the primary outcome measurement was binary classification (Yes [1] vs No [0]), we explored if LLMs could apply more complex measurement metrics and how this approach might differ from using binary classification. We applied the PHQ-4, a simplified yet validated diagnostic tool for depression and anxiety.[22] Employing systemic persona and zero-shot learning, we directed LLMs to do multiple tasks, involving multi-class and multi-level classification to: 1) assess each message for four categories (little interest, hopelessness, nervousness, worrying); 2) rate the message for four levels using a 4-point Likert scale (0=not at all; 3=most likely) in each category; 3) calculate the sum of all the ratings; and 4) categorize it into No [0] if sum < 6, or Yes [1] if sum ≥ 6, applying the original scoring standard of PHQ-4. We performed z tests to compare the performance of binary and PHQ-4 based classifications by model. The statistical significance was determined at $p<0.05$.

**With vs without systemic persona**. To enhance LLMs' performance, we meticulously crafted a systemic persona using multiple prompting strategies[23,24]: 1) role prompting (e.g. as Dr. GPT, a professional psychiatrist, your role is~), 2) directive commanding (e.g. evaluate the message~, be sure to offer~), 3) expertise emulation (e.g. I myself am a psychiatrist in the hospital), 4) zero-shot chain of thought (e.g. take time to think deeply and step-by-step to be sure). The full-engineered prompts are in **Appendix 3**.

*Exploration of LLMs using challenging cases*
We further assessed the performance of LLMs using challenging cases ($N$=39). Challenging positive benchmark (Yes [1]$_{challenges}$, $n$=14) comprised the messages with symptoms that required further discussion by two experienced psychiatrists (CIR and PJR), because they involved differential diagnosis and proxy symptoms (e.g. demoralization, anticipatory anxiety). If two psychiatrists agreed that they would flag the patient for further assessment, it was labeled as positive benchmark. Challenging negative benchmark (No [0]$_{challenges}$, $n$=25) comprised the messages that contained at least one mental health keyword, but were determined as not depressive or anxious by the two psychiatrists.

We aimed to see if LLMs could understand the overall context, not relying heavily on the appearance of the signaling words to detect symptoms. In this post hoc assessment, we additionally explored the performance of two of the latest reasoning models (OpenAI o3-mini [OpenAI Inc., January 31, 2025] and Gemini Pro 2.0 Thinking model [Google LLC., January 21, 2025]), hypothesizing that the reasoning models may have higher performance in challenging cases. We assessed the performance of binary classification of six LLMs regarding F-1 score, precision, recall, and accuracy in zero-shot settings. At this time, LLMs were required to provide their reasoning along with classification. Two experienced psychiatrists (CIR and PJR) reviewed the LLMs' classification reasoning to detect any unreasonable rationale or hallucinations. LLMs' reasoning and classification are in **Appendix 4**.

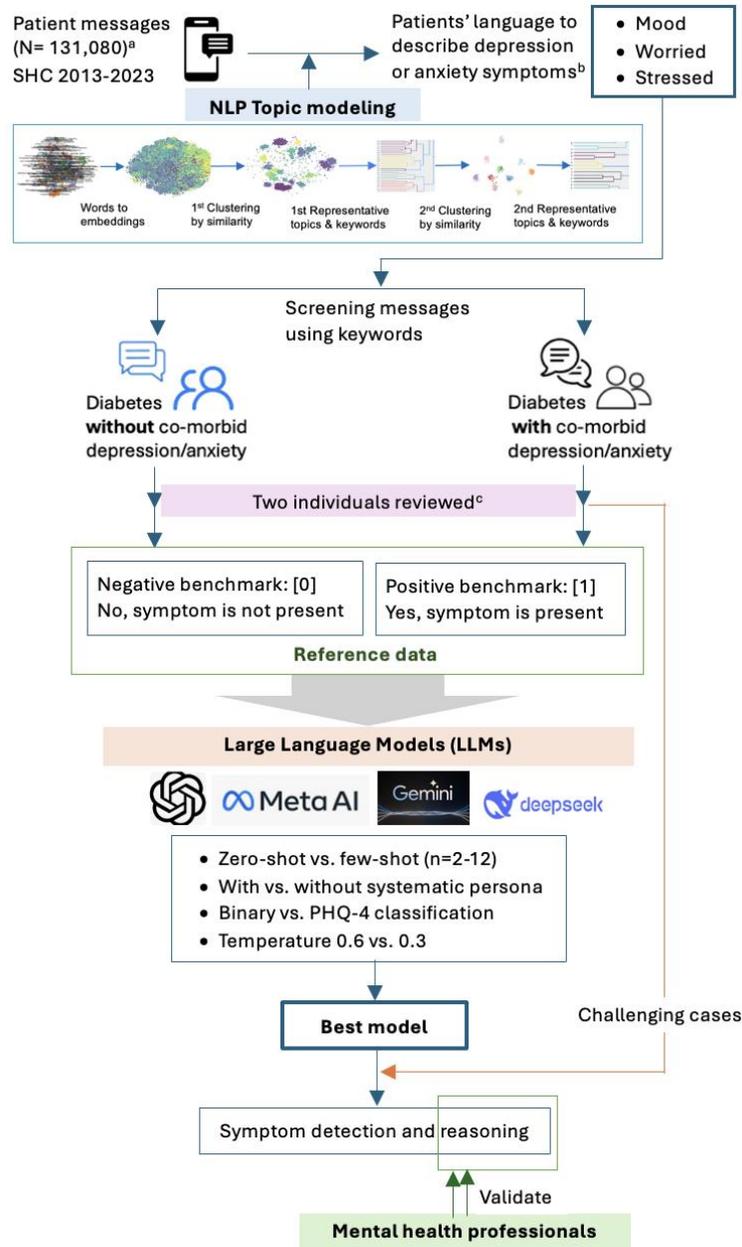

**Figure 1. Data sources and study flow**

[a] Messages from those with diabetes and co-morbid depression or anxiety; [b] Symptomatic keywords grouped with depression or anxiety were extracted; [c] a health researcher reviewed, and a psychiatrist confirmed applying predefined criteria. Only the concordant messages between two reviewers were included as the benchmark

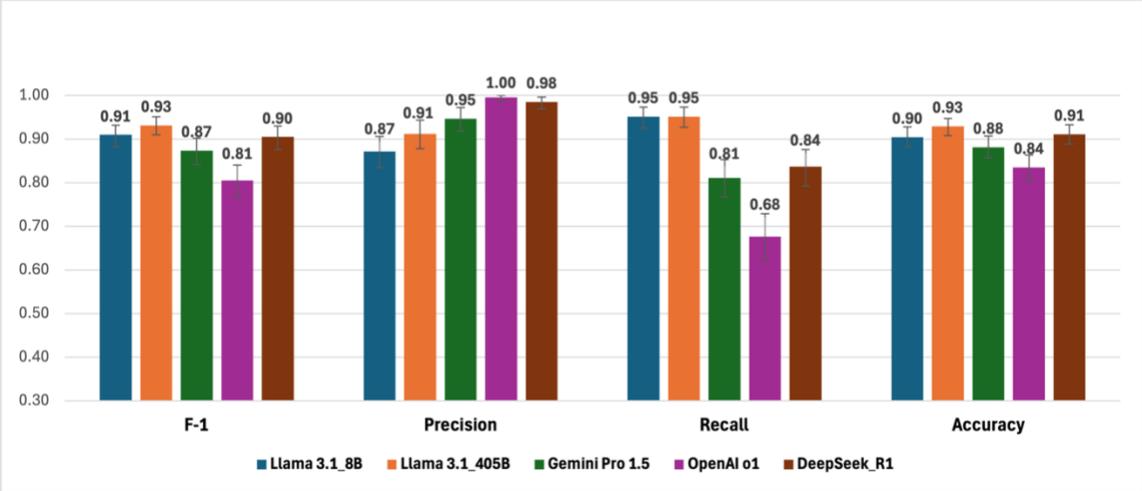

a) Zero-shot (F-1, Precision, Recall, and Accuracy)

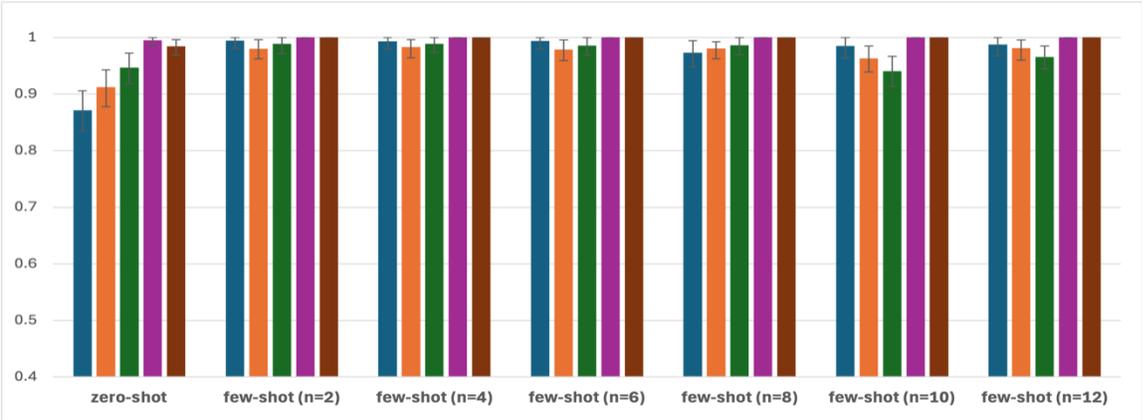

b) Few-shot (Precision)

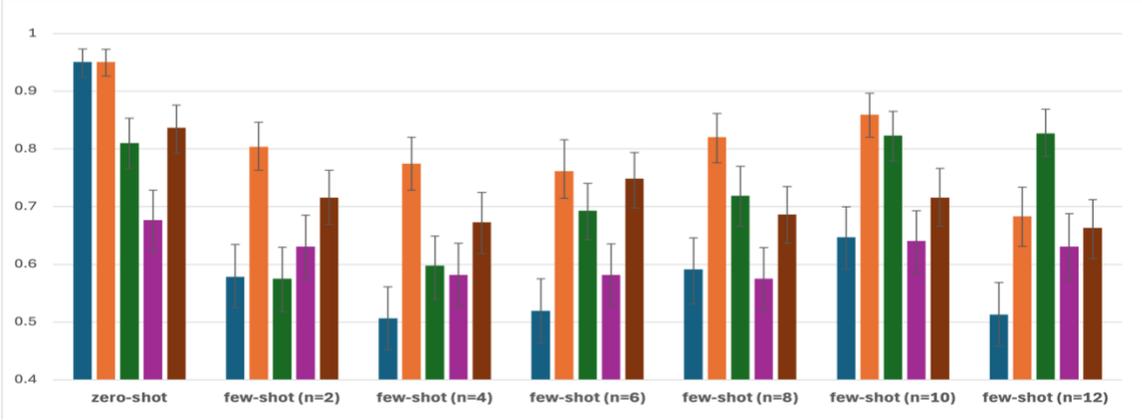

c) Few-shot (Recall)

**Figure 2. Performance of LLMs: Zero-Shot vs Few-Shot**

All results (F-1, precision, recall, and accuracy for *N*=606) of zero-shot vs few-shot by model are available in Appendix 5.

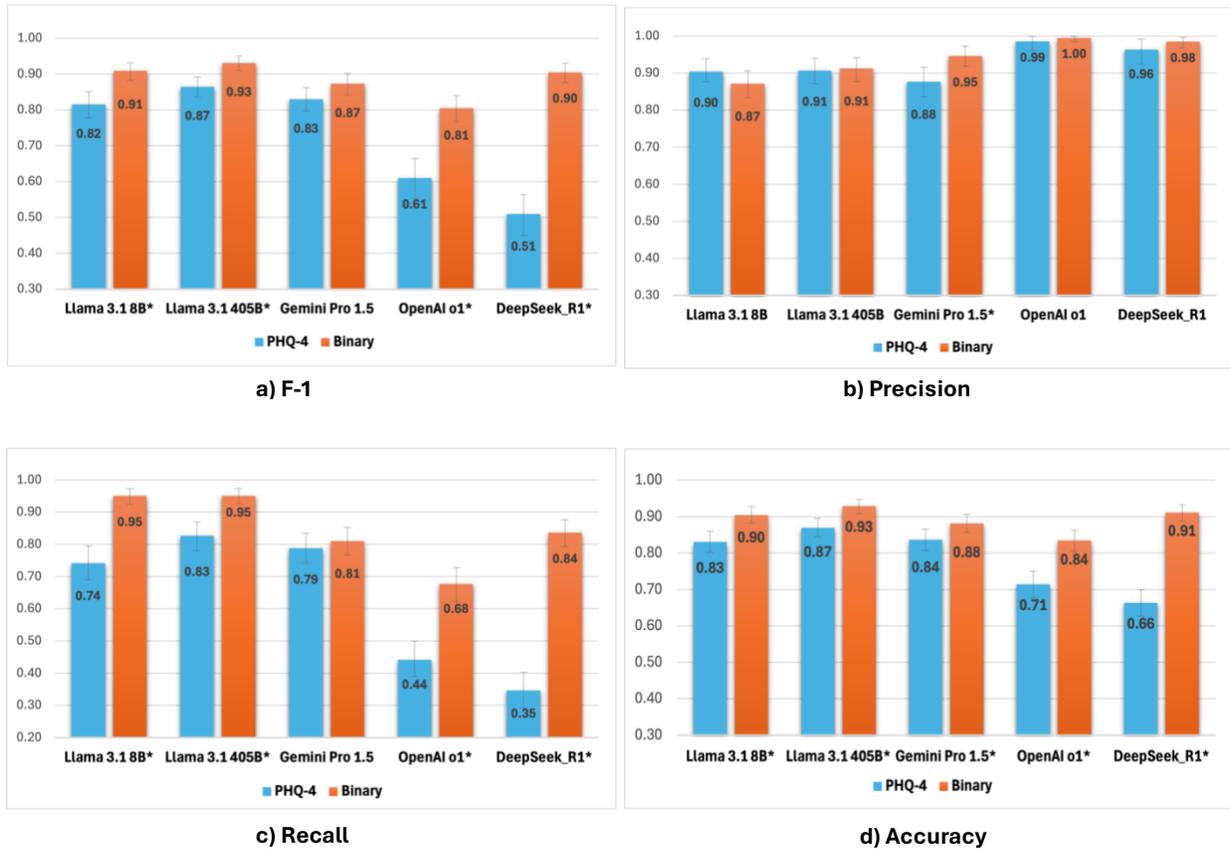

**Figure 3. Performance of LLMs: Binary vs. PHQ-4 classification**

 * Statistically different by z-test (p<0.05); Binary: Yes [1] vs No [0]; PHQ-4: i) assess each message for four categories (little interest, hopelessness, nervousness, worrying), ii) rate the message for four levels using a 4-point Likert scale (0=not at all; 3=most likely) in each category, iii) calculate the sum of all the ratings, iv) categorize it into No [0] if sum < 6, or Yes [1] if sum ≥ 6.

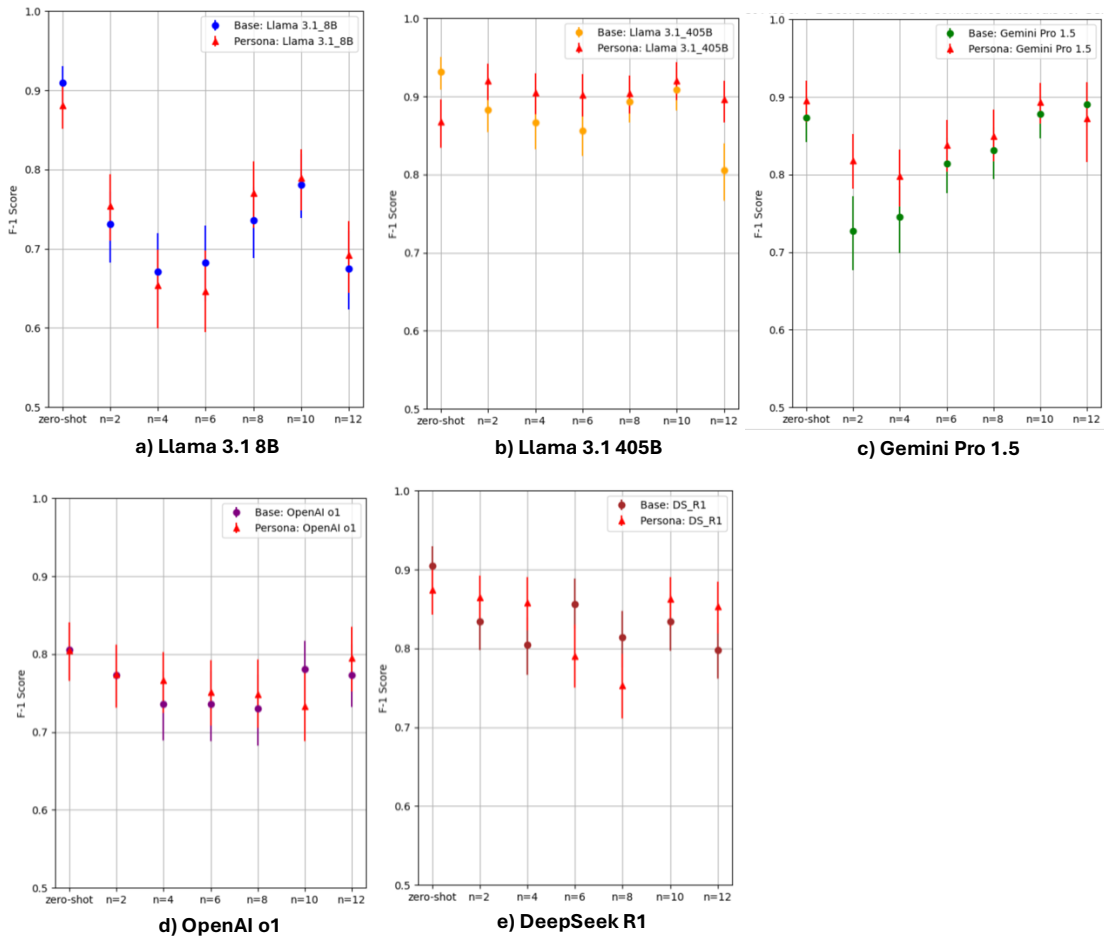

**Figure 4. Effect of Systemic Persona on F-1 Score by LLM**

▲ With systemic persona ; *N*=606; All results (F-1, precision, recall, and accuracy) of the effect of systemic persona by model are available in Appendix 6.

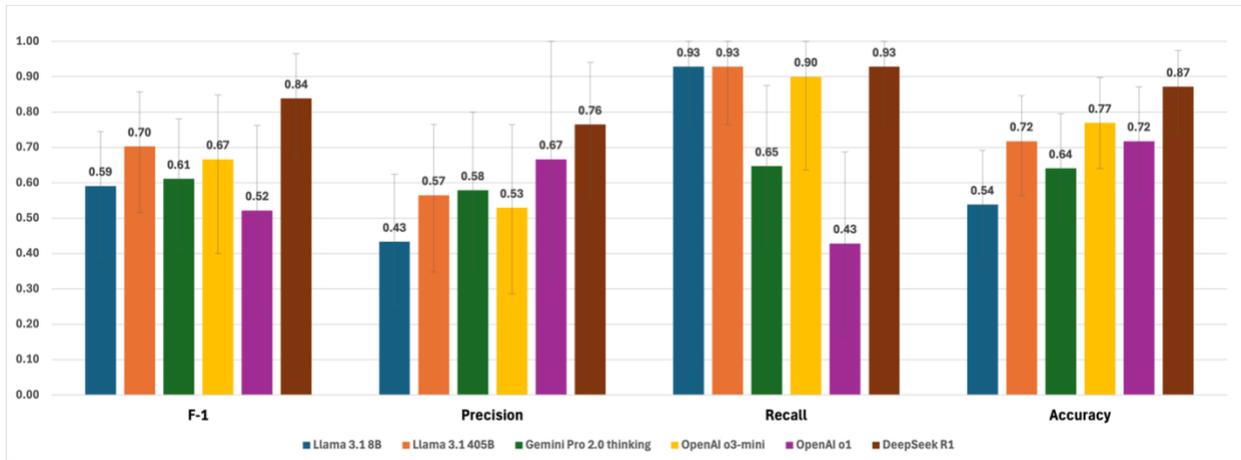

**Figure 5. Performance of LLMs in Challenging Cases**
Challenging cases (*N*=39); LLMs' reasoning and classification are in Appendix 4.